\documentclass[10pt,conference,a4paper]{IEEEtran}
\IEEEoverridecommandlockouts
\usepackage{cite}
\usepackage{amsmath,amssymb,amsfonts}
\usepackage{algorithmic}
\usepackage{graphicx}
\usepackage{textcomp}
\usepackage{xcolor}
\def\BibTeX{{\rm B\kern-.05em{\sc i\kern-.025em b}\kern-.08em
    T\kern-.1667em\lower.7ex\hbox{E}\kern-.125emX}}

\usepackage{times}
\usepackage{epsfig}
\usepackage[ruled,vlined]{algorithm2e}
\usepackage{gensymb}
\usepackage{stfloats}
\usepackage{booktabs}
\usepackage{threeparttable}
\usepackage{multicol}
\usepackage{multirow}
\usepackage{subfigure}
\usepackage{color}
\usepackage{caption}
\usepackage{lipsum}
\usepackage{float}
\usepackage{setspace}

\usepackage[font=small,skip=8pt]{caption}
\usepackage[colorlinks,
            linkcolor=green,
            anchorcolor=green,
            citecolor=green]{hyperref}

\usepackage{colortbl}
\definecolor{mygray}{gray}{.9}
\definecolor{mypink}{rgb}{.99,.91,.95}
\definecolor{mycyan}{cmyk}{.3,0,0,0}

{}
{}
{}
\DeclareRobustCommand*{\IEEEauthorrefmark}[1]{%
    \raisebox{0pt}[0pt][0pt]{\textsuperscript{\footnotesize\ensuremath{#1}}}}

\begin{document}

\title{RSINet: Rotation-Scale Invariant Network for Online Visual Tracking}

\author{\IEEEauthorblockN{Yang Fang\IEEEauthorrefmark{1}\textsuperscript{,}\IEEEauthorrefmark{2}, Geun-Sik Jo \IEEEauthorrefmark{2}\textsuperscript{*}\thanks{* Geun-Sik Jo is the corresponding author (e-mail: gsjo@inha.ac.kr).}, Chang-Hee Lee\IEEEauthorrefmark{1}, \textit{Fellow, IEEE}}
\IEEEauthorblockA{\IEEEauthorrefmark{1}Dept. of Electrical Engineering, Korea Advanced Institute of Science and Technology (KAIST), Daejeon, Republic of Korea}
\IEEEauthorblockA{\IEEEauthorrefmark{2}School of Computer and Information Engineering, Inha University, Incheon, South Korea}
Email: fangyang@kaist.ac.kr, gsjo@inha.ac.kr, changheelee@kaist.ac.kr
}
\maketitle

\begin{abstract}
Most Siamese network-based trackers perform the tracking process without model update, and cannot learn target-specific variation adaptively. Moreover, Siamese-based trackers infer the new state of tracked objects by generating axis-aligned bounding boxes, which contain extra background noise, and are unable to accurately estimate the rotation and scale transformation of moving objects, thus potentially reducing tracking performance.
In this paper, we propose a novel Rotation-Scale Invariant Network (RSINet) to address the above problem. Our RSINet tracker consists of a target-distractor discrimination branch and a rotation-scale estimation branch, the rotation and scale knowledge can be explicitly learned by a multi-task learning method in an end-to-end manner. In addtion, the tracking model is adaptively optimized and updated under spatio-temporal energy control, which ensures model stability and reliability, as well as high tracking efficiency. Comprehensive experiments on OTB-100, VOT2018, and LaSOT benchmarks demonstrate that our proposed RSINet tracker yields new state-of-the-art performance compared with recent trackers, while running at real-time speed about 45 FPS.
\end{abstract}
\begin{IEEEkeywords}
Siamese tracker, rotation-scale invariance, spatio-temporal consistency, real-time tracking
\end{IEEEkeywords}

\section{Introduction}
Visual object tracking is a challenging but important problem in the computer vision field, which aims to predict the new state of the tracked object in unseen frames given an initial bounding box annotation at the first frame. It plays an essential role in many real-world applications, such as public surveillance and security, human-computer interaction, autonomous driving systems and robotic services. Although object tracking has been studied for several decades, and much progress has been made in recent works \cite{KCF, DSST, CCOT, ECO, MDNet, SINT, Siamese-FC, GLST, GOUTURN, SiameseRPN}, it remains a difficult problem because of the many challenging aspects \cite{OTB-100, VOT-2018} in visual object tracking, such as illumination variation, scale variation, in-plane and out-of-plane rotations, etc.
\begin{figure}[hbt]
	\centering
	\includegraphics[width= 0.45\textwidth]{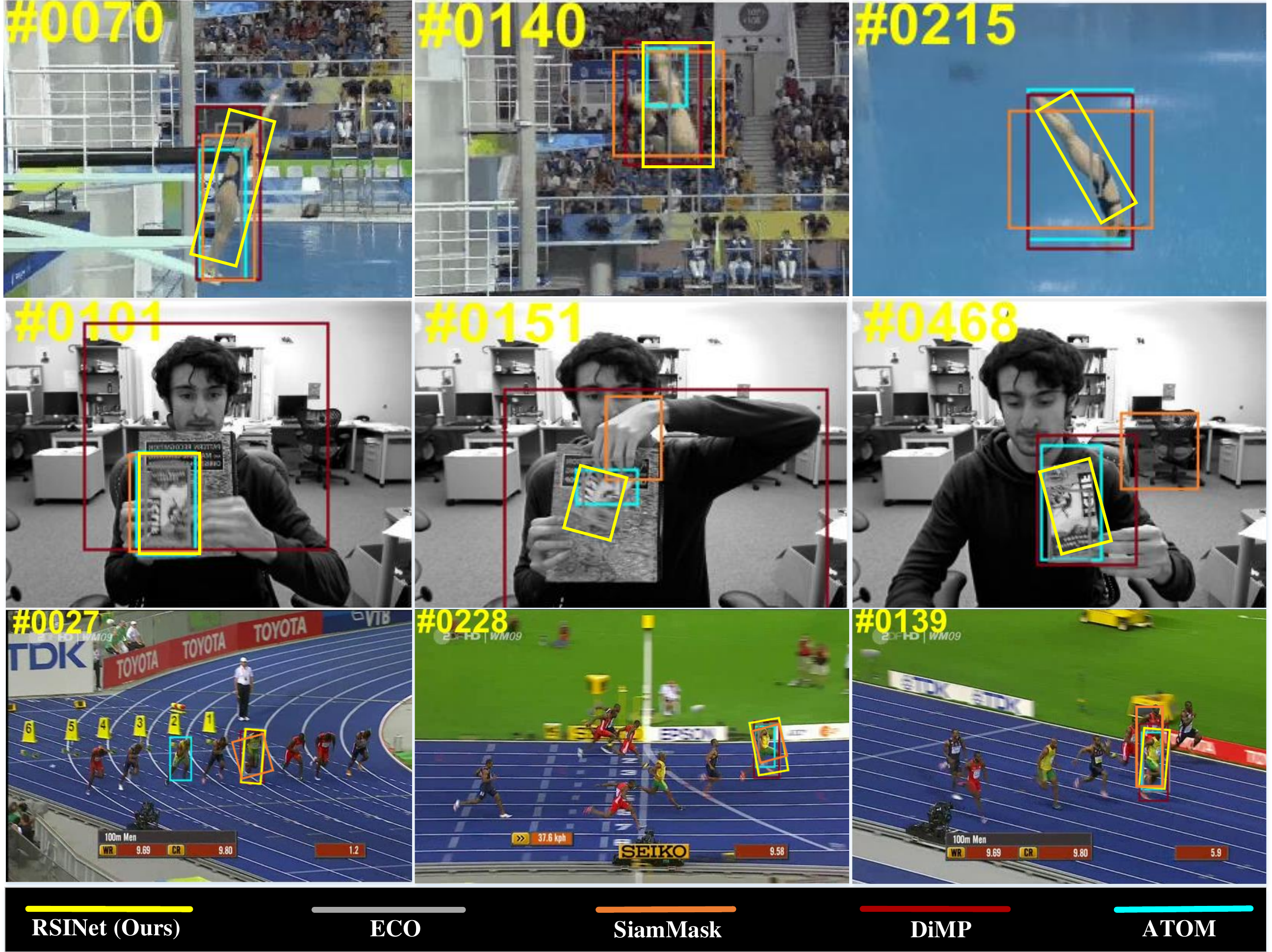}
	\caption{Tracking results visualization generated by our proposed RSINet tracker and four recent SOTA trackers, ECO \cite{ECO}, SiamMask \cite{SiamMask}, DiMP \cite{DiMP}, and ATOM \cite{ATOM}. RSINet tracker performs better in object rotation and scale variance circumstances than other trackers.}
	\label{fig:tracking_visualization1}
\end{figure}
Correlation filter-based trackers have shown excellent tracking performance in terms of both tracking accuracy and speed due to the efficient circular sampling method and fast Fourier transform. Henriques {\em et al}. \cite{KCF} propose the Kernelized Correlation Filter (KCF), which extends the linear correlation filter \cite{MOSSE, CF-2012} by applying a kernel trick while having the same complexity as linear regression. Danelljan {\em et al}. make a milestone contribution to the development of a Discriminative Correlation Filter (DCF) for visual object tracking. They propose a series of DCF-based tracker variants \cite{CNTracker, SRDCF, DSST, CCOT, ECO}. DSST tracker \cite{DSST} builds two correlation filter branches and performs target position estimation and scale estimation iteratively, which increases tracking accuracy at the expense of running speed. From continuous convolution operators tracker (CCOT) \cite{CCOT}, they enhance the DCFs by introducing a continuous correlation operator. Moreover, efficient convolution operators (ECO) tracker \cite{ECO} applies adaptive feature map compression and a Gaussian Mixture Model to improve tracking accuracy while simultaneously increasing the tracking speed. An Aberrance Repressed Correlation Filter (ARCF) \cite{ARCF} suppresses aberrancies during the target detection phase by incorporating a restriction to the rate of alteration with response maps, thereby achieving a new state-of-the-art performance in aerial-view object tracking tasks. Lu {\em et al}. \cite{RPCF} firstly introduce region-of-interest (ROI) pooling in correlation filters for visual tracking, which compresses the model while preserving the localization accuracy to yield a new state-of-the-art on many tracking benchmarks.

Recently, Siamese network-based trackers \cite{CFNet, SiamRPN++, DaSiamese, SiamMask, DWSiamese} have attracted strong attention due to their high tracking speed and accuracy. Siamese trackers consider the visual tracking problem by constructing a similarity map between the target template and the search region. Most Siamese trackers are trained on large-scale sequence data, and during the tracking process they apply an offline learned model to predict new object locations while freezing the network parameters to ensure high tracking efficiency. Specifically, Bertinetto {\em et al}. \cite{Siamese-FC} first introduce an end-to-end fully-convolutional Siamese network (SiameseFC) for visual tracking tasks. This network formulates tracking as a process that matches the target template with sliding-window patches of the search region by cross-correlation operation, and the image patch with the maximum matching score is selected as the new target object. Since the SiameseFC tracker applies a lightweight network as the encoder and uses a simple cross-correlation model for prediction, it lacks the ability to discriminate the target object from background distractors. CFNet \cite{CFNet} incorporates a Correlation Filter layer into the fully-convolutional Siamese framework, which can online learn the parameters of the Correlation Filter layer and help improve the tracking performance of only lightweight networks but not deep network-based trackers.
Moreover, most non-RPN Siamese trackers \cite{SINT, Siamese-FC, GOUTURN, CFNet} focus on target location estimation without using a specific model for scale or rotation estimation, whereas RPN-based Siamese trackers \cite{SiameseRPN, SiamRPN++, DaSiamese, SiamMask} adopt a predefined anchor box of the region proposal network for final axis-aligned bounding boxes regression, which cleverly eliminates the scale estimation process; however object rotation information learning remains absent. Wang {\em et al}. \cite{SiamMask} firstly introduce a binary segmentation mask for generating rotated object bounding boxes along with heavy optimization cost, and it is vulnerable in non-object target tracking.

In this paper, to overcome the above restrictions, we propose a novel Rotation-Scale Invariant Network (RSINet) to make rotation and scale variance learnable in a Siamese-based learning framework. Different from the current Siamese trackers, our RSINet enables object-aligned target tracking and adaptive model update in an highly-efficient end-to-end manner.
We summarize the three main contributions of this work as follows:

\begin{itemize}

\item We propose a unifying framework consisting of a target-distractor discrimination branch and a scale-rotation estimation branch, which can generate a more precise target representation than the existing state-of-the-art Siamese trackers.

\item To naturally predict target state, including target location, rotation and scale variance, we apply a Siamese-based network to learn the task-specific but shared feature representation for multi-task leaning, given input from different sampling space. And we propose an adaptive model update approach under spatio-temporal energy control for model stability and reliability during tracking process.

\item Our RSINet yields a new state-of-the-art performance while running at high tracking speed demonstrated with comprehensive experiments on OTB-100, VOT2018 and large-scale LaSOT datasets.

\end{itemize}

The rest of this paper is structured as follows. Section \ref{RelatedwWork} introduces most of the relevant prior works with this paper; Section \ref{RSINet} presents the proposed RSINet model in detail; Section \ref{experiments} describes the experimental results and the analysis of the RSINet tracker; Section \ref{conclusion} draws the conclusion for this paper.

\section{Related Work} \label{RelatedwWork}
In this section, we briefly review the following works related with our proposed tracking framework: Siamese-based trackers, the online discriminative learning model and Log-Polar coordinate transform-based trackers.

 {\bf Offline Siamese trackers.} Li {\em et al} \cite{SiameseRPN} first propose the Siamese region proposal network (SiameseRPN) to efficiently solve the scale variation estimation challenge by applying an advanced region proposal network (RPN) together with Siamese architecture. The SiameseFC tracker can predict much more accurate bounding boxes and improve the tracking accuracy compared with previous non-RPN-based Siamese trackers. SiamRPN++ \cite{SiameseRPN} and DWSiamese \cite{DWSiamese}, which are enhanced versions of original SiameseRPN, replace shallow backbone network of SiameseRPN (such as AlexNet \cite{AlexNet} and VGGNet \cite{VGG}) with very deep network architecture (ResNet \cite{ResNet} and Inception \cite{Inception}) to further improve the tracking performance of their baseline. DaSiamese \cite{DaSiamese} tries to overcome the imbalanced distribution of the training data by learning a distractor-aware module to enhance the discriminative power of learned features. Wang {\em et al}. \cite{SiamMask} propose a SiamMask network to integrate both visual object tracking and video object segmentation in a unifying framework. It firstly generates target-aligned instead of axis-aligned bounding boxes for more accurate object representations, thus making a new state-of-the-art among previous Siamese trackers. However, their target representations are optimized from a separate binary mask branch, which substantially determines the quality of the target bounding box but without learning real rotation and scale information, and its heavy optimization strategy leads to low tracking speed.

 {\bf Online deep trackers.} Danerlljan {\em et al} propose an ATOM tracker \cite{ATOM} to focus on accurate state estimation by target-specific-based overlap maximization with an online classification component, which is different from most offline Siamese trackers that do not online learn target-specific knowledge and lack the ability to integrate distractor information. In another of their works \cite{DiMP}, they further emphasize the nature of the visual tracking problem that requires online learning of a target-specific appearance model in the tracking stage. They develop an end-to-end tracking framework, capable of fully learning the target and background appearance model for target model inference, and it achieves a new state-of-the-art with a real-time running speed. Similar with current Siamese trackers, DiMP does not learn the rotation and scale information, which is the main limitation for its tracking accuracy.

 {\bf Log-Polar-based coordinate based trackers.} To explicitly learn scale and rotation information, Li {\em et al}. \cite{SRCF} propose a scale-and-rotation correlation filter (SRCF) that formulates the scale and rotation between two images in pure Log-Polar coordinate translations with Fourier domain transform, and its performance goes beyond traditional correlation filter trackers. Zokai {\em et al}. \cite{Zokai} propose a hybrid model that integrates log-polar mappings and nonlinear least squares optimization for recovering large similarity transformation. Li {\em et al}. \cite{LDES} apply a log-polar-based phase correlation approach with an efficient coordinates descent optimizer to estimate both rotation and scale variance simultaneously. However, they all apply only hand-crafted features (such as histogram of orientation gradient) for target appearance representations that are very limited for target representation and robust discrimination model learning.

 Instead, we apply Siamese-based CNN network to extract the task-specific but shared feature representation for target location, rotation and scale estimation in Cartier and Log-polar sampling space simultaneously. For target tracking and model learning efficiency, we apply an adaptive model update method to optimize the discrimination branch and scale-rotation invariance branch jointly in real-time. To the best of our knowledge, this is the first work to model the online object-aligned tracking task by applying different sampling space data to learn a shared CNN network in an end-to-end manner.

\begin{figure*}
\centering
\includegraphics[width= 0.8\textwidth]{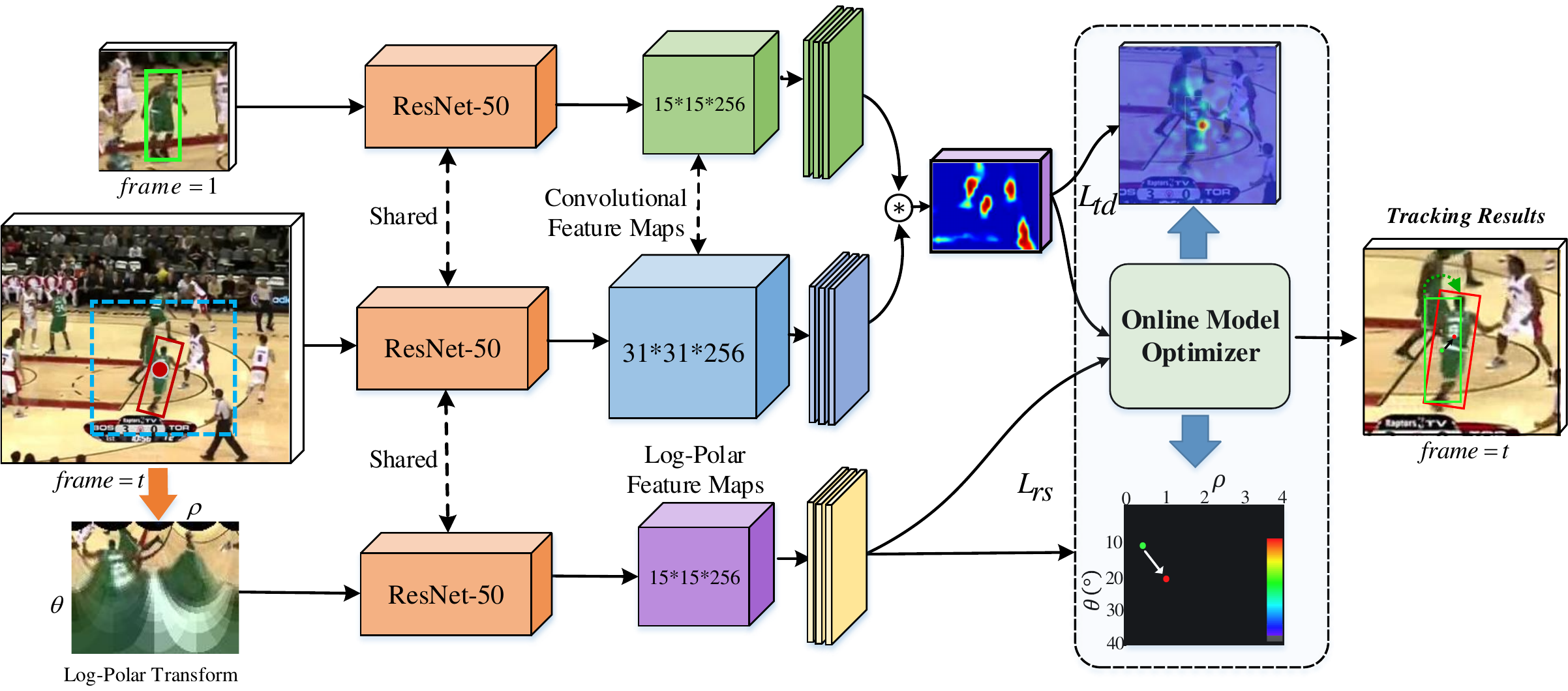}
\caption{Our proposed rotation-scale invariant network (RSINet) architecture. It consists of two main branches: target-distractor discrimination module (top) and rotation-scale invariant module (bottom). Both are learned and updated by online model optimizer during training and testing stages. RSINet tracker predicts target position, rotation angle and scale factor simultaneously at each frame. }
\label{fig:RSINet-framework}
\end{figure*}
\section{Rotation-Scale Invariant Network (RSINet)} \label{RSINet}
In this section, we first propose a novel tracking framework consisting of two components: \ref{RSImodule} A rotation-scale Invariance (RSI) module that explicitly exploits the displacement of rotation and scale in the Log-Polar coordinate, and \ref{TDDmodule} A target-distractor discrimination (TDD) module, which is dedicated for target position estimation. Both of them are jointly learned by an efficient online optimizer during the {\em online} tracking stage. The network training details are presented in \ref{networktraining}. And online tracking and adaptive model update are addressed in \ref{modelupdate}.

\subsection{Rotation-Scale Invariance Module}\label{RSImodule}
To explicitly estimate the change of the scale factor and rotation angle between different frames, we propose a light-weight rotation-scale invariant deep feature extraction network that generates deep feature maps of a Log-Polar (LP) target template in the Log-Polar coordinate system. Given an anchor image $I(x,y)$ with spatial position $(x,y)$ in the Cartesian coordinate system, it can be transferred into Log-Polar coordinates by a nonlinear and nonuniform function T as $\hat{I}(\rho,\theta) = T(I(x,y))$. Similar with polar coordinates, the Log-Polar coordinate system has a common origin point and two axis directions. The logarithm of the distance between a point and the pole is aligned with the horizontal axis represented with $\rho$, while the relative rotation angle between one point and the polar axis is aligned with the vertical axis represented by $\theta$.
Specifically, let $I_{t}(x,y)$ represent the tracked target object in frame $t$, and  $I_{t+1}(x,y)$ represent the tracked target object in frame $t+1$ with spatial translation ($\Delta x$, $\Delta y$), rotation angle $\theta$ and scale factor $\rho$; the spatial translational relationship between $I_{t}(x,y)$ and $I_{t+1}(x,y)$ is formulated as
\begin{equation}\small
\begin{split}
I_{t+1}(x,y) = I_{t}(&\frac{x\cos(\theta) + y \sin(\theta) - \Delta x}{\rho}, \\
&\frac{-x \sin(\theta) + y\cos(\theta) - \Delta y}{\rho})
\end{split}
\label{LPT}
\end{equation}
We use the Fourier magnitude spectra to approximate Eq. (\ref{LPT}); the relationship between $I_{t}$ and $I_{t+1}$ can be rewritten as
\begin{equation}\small
\begin{split}
\mathcal{I}_{t+1}(\mathtt{x},\mathtt{y}) = \mathcal{I}_{t}(&\frac{\mathtt{x}\cos(\theta) + \mathtt{y} \sin(\theta)}{\rho}, \\
&\frac{-\mathtt{x} \sin(\theta) + \mathtt{y}\cos(\theta)}{\rho})
\end{split}
\label{LPTF}
\end{equation}
Where $\mathcal{I}$ is the Fourier transform of $I$, $\mathtt{x}$ and $\mathtt{y}$ are the Fourier transform of $x$ and $y$, respectively. By Fourier transform analysis and approximation, the spatial translation ($\Delta x$, $\Delta y$) is omitted and only the rotation angle $\theta$ and scale factor $\rho$ are preserved in Eq. (\ref{LPTF}). Therefore, we consider the change of target rotation and scale between $I_{t+1}$ and $I_{t}$ as $I_{t}(e^{\rho}\cos(\theta),e^{\rho}\sin(\theta)) = I_{t+1}(e^{\rho + \Delta \rho}\cos(\theta + \Delta \theta),e^{\rho + \Delta \rho}\sin(\theta + \Delta \theta))$ in Cartesian coordinates. Therefore, it can be derived in Log-Polar coordinates as the following formula
\begin{equation}\small
\begin{split}
I_{t+1}^{lp}(\rho, \theta) = I_{t}^{lp}(\rho - \Delta \rho, \theta - \Delta \theta)
\end{split}
\label{LPRelation}
\end{equation}

Instead of using hand-crafted feature representations of the Log-Polar target image $I^{lp}$ and phase-correlation filter learning \cite{LDES}, in this paper we apply pretrained ResNet-50 network to extract the deep {\em Conv} features maps as rotation-scale invariance features and use a three-layer fully convolutional network as Log-Polar based rotation-scale translation regressor in a data-driven manner. The original target patch $I$ is firstly transformed to Log-Polar image $I^{lp}$, and then $I^{lp}$ is fed forward into feature extraction network and three sequential fully-convolutional regressor. Each of fully-convolutional layer is followed by an activate function $\{\psi_{i}\}_{i =1}^{3}$; the network architecture is formulated as
\begin{equation}\small
\begin{split}
f(I^{lp}, {\bf h}) &= \psi_{3}(h_{3}\ast \psi_{2}(h_{2}\ast \psi_{1}(h_{1} \ast I^{lp})))\\
& = (\rho^{\star}, \theta^{\star})
\end{split}
\label{three-fc}
\end{equation}
The last convolutional layer has two head branches that directly predict the Log-Polar coordinates ($\rho^{\star}$, $\theta^{\star}$). During the training stage, the network accepts a pair of Log-Polar training samples, $S_{t} = \{I_{t}^{lp}, I_{t+\tau}^{lp}, (\rho_{t}, \theta_{t})\}$, as the network input. Here $I_{t}^{lp}$ acts as the anchor image sampled at frame $t$ of the training sequence and $I_{t+\tau}^{lp}$ is the rotated and scaled translation of $(\rho_{t}, \theta_{t})\}$ w.r.t the anchor image $I_{t}^{lp}$ that is sampled with frame interval $\tau$. The final target-distractor discrimination loss is built as
\begin{equation}
\begin{split}
L_{rs}({\bf h}) = \sum_{i=1}^{N} \Arrowvert \mathcal{R}(f(I_{i}^{lp}, h), g_{i})\Arrowvert^{2} + \sum_{j} \lambda_{j} \Arrowvert h_{i} \Arrowvert^{2}
\end{split}
\label{L-td}
\end{equation}
Where $g_{i} = (\rho,\theta)$ is the ground-truth label of rotation and scale translation between the anchor image $I_{t}^{lp}$ and the translated sample $I_{t+\tau}^{lp}$. The $\mathcal{R}$ is the final residual that is back-propagated to the network for optimization; it is defined as
\begin{equation}
\begin{split}
\mathcal{R}(f(I_{i}^{lp}, h), g_{i}) &= (\frac{\rho^{\star} - \rho}{\rho},\frac{\theta^{\star} - \theta}{\theta})\\
& = (\Delta \rho, \Delta \theta)
\end{split}
\end{equation}
Both the rotation residual $\Delta \rho$ and scale residual $\Delta \theta$ are normalized by dividing them with ground-truth translation to guarantee the loss balance during the training process. The detailed training approach and dataset collection are presented in section \ref{offline-training} and the online learning details are presented in section \ref{online-learning}.

\subsection{Target-Distractor Discrimination Module}\label{TDDmodule}
In this section, we detail the construction of the target-distractor discrimination module that is inspired by the state-of-the-art online deep tracker \cite{ATOM, DiMP}. The aim of the discrimination module is to learn the discriminative filters of convolutional layers by a dedicated discriminative loss function that can learn not only target-specific information but also the background distractors near the target. Specifically, the discriminative module M accepts training samples $S = {(x_{i}, y_{i})}_{i = 1}^{N}$ as network input, where $x_{i}$ stands for the deep feature representations extracted from original training images by the shared ResNet50 backbone network \cite{ResNet}, and $y_{i}\in [0,1]$ is the corresponding ground-truth labels  of $x_{i}$, usually set to a Gaussian function that assigns the highest label value for target center position and near-zero value in the background region. Given these training data pairs, our discriminative module aims to learn a target-distractor aware model composed of convolutional weights. Least-squares regression-based discriminative loss function has achieved great success in correlation filter trackers \cite{DSST, ECO, Context-Aware-CF} thanks to its advantages of sample implementation form and efficient optimization mechanism, which can be formulated as the following formula,
\begin{equation}
L_{td}({\bf w}) = \frac{1}{N} \sum_{(x, y) \in S}\Arrowvert s(x,w) - y\Arrowvert^{2} + \Arrowvert \gamma \ast w \Arrowvert^{2}
\end{equation}
However, through experimental analysis we find that this conventional discriminative loss focuses more on regressing negative samples score to zero instead of learning the discriminative properties between positive and negative samples due to the imbalanced distribution of the training data set. To mitigate this negative impact, we follow the strategy of support vector machine (SVM) \cite{SVN, DiMP} and build a hinge-like distractor-aware discriminative score map formula $s(x, w) = m \cdot (x \ast w) + (1-m)\cdot \max(0,x\ast w)$,
the modified score map formula can masterly refrain from learning the residual of obvious negative training samples and can focus solely on enhancing the discriminating ability. Finally, our end-to-end tracking framework facilitates the learning of the convolutional filters $w$, the regularization factor $\gamma$, and the score map mask $m$ jointly, during both offline and online tracking stages. Note that the RSI module and TDD module share same backbone ResNet-50 network for extracting common feature maps as shown in Fig. \ref{fig:RSINet-framework}. We detail the training process and online learning method in section \ref{offline-training} and section \ref{online-learning}, respectively.

\subsection{Network Training Details}\label{networktraining}
\label{offline-training}
In this section, we mainly present the training approach of the rotation-scale invariance module. We collect training data from several large-scale annotated video datasets, including ILSVRC2017 VID \cite{ILSVRC2017}, MS COCO \cite{COCO}, LaSOT \cite{LaSOT} and GOT10k \cite{GOT10k}. To build training image pairs $(I_{t}, I_{t+\tau})$ for training the rotation-scale invariance module, we randomly sample rotated and scaled image patches $\{I_{t+\tau}\}_{\tau = 1}^{M}$ for each annotated sample $I_{t}$. Assuming that the maximum value of the absolute rotation angle $\theta$ between two constructive frames is not exceeded $30^{\circ}$, and we sample ten different rotation degrees as $\{(3\times d)^{\circ}\}_{d=1}^{10}$. The scale factors $\rho$ between two constructive frames are within the range $(\frac{1}{1.1^{5}}\leq \rho \leq 1.1^{5})$, we sample the scale factor as $\{\frac{1}{1.1^{5}}, \frac{1}{1.1^{4}}, \dots, 1.1, \dots, 1.1^{4}, 1.1^{5}\}$, that are total 10 different scale factors. By meshing the rotation angles and scale factors in a meshgrid matrix, we can generate total $M =100$ $(10\times 10)$ rotation-scale samples $\{I_{t+\tau}\}_{\tau =1}^{100}$ w.r.t the anchor image $I_{t}$. Then the training sample pairs are built as $S_{t} = \{I_{t}^{lp}, I_{t+\tau}^{lp}, (\rho_{t}^{\tau}, \theta_{t}^{\tau})\}$ for objective loss calculation. We firstly divide each video sequence into video segments, with each segment containing 15 sequential frames, and the middle ($7th$) frame is selected as an anchor image. We sample the rotation-scale translation on each anchor image and totally construct around 400,000 training sample sets. We set the mini-batch size to $N = 64$ to calculate the rotation-scale invariance module loss $L_{rs}$.

For target-distractor discrimination module training, we follow the training method and training data collection in \cite{DiMP} to calculate the target-distractor discrimination loss $L_{td}$. Instead of applying IoU-Net \cite{IoU-Net} for bounding box estimation \cite{ATOM}, we apply our proposed RSINet for rotation and scale aware bounding box generation. The final training loss function is defined as the weighted summation of RSI loss $L_{rs}$ and TDD loss $L_{td}$ with a trade-off factor $\mu$,
\begin{equation}
L = L_{rs} + \mu L_{td}
\end{equation}
During tracking process, the target-disctractor discrimination module is firstly executed for target position prediction, and the rotation-scale invariant module is then performed for rotation and scale estimation based on predicted target position. And reliable position prediction plays a crucial role for accurate tracking performance, therefore we set $\mu = 50$ to ensure that our network to pay more attention for target-distractor discrimination learning during training process.
We train RSINet for 60 epoches with 15000 videos per epoch on a single Nvidia TITIAN X GPU for 40 hours. We apply the ADAM \cite{ADAM} optimizer with learning rate decay of 0.2 every 10th epoch.

\subsection{Online Tracking and Adaptive Model Update}\label{modelupdate}
\label{online-learning}
The pre-trained RSINet network performs as an online object tracker, it accepts an initial axis-aligned target annotation as input, and the shared backbone network extracts deep feature maps and feeds forward them into feature calibration block to generate domain-specific features for RSI and TDD modules, respectively. The discrimination module applies domain-specific features for target position prediction, while rotation-scale invariant module applies both shared and task-specific feature representation for angular rotation and scale variance estimation. To make tracking model adapt to change in target appearance and scenario, tracking model needs to be updated accordingly during tracking process. Instead of applying conjugate gradient (CG) \cite{ECO} optimization method with intuitive update rate, we propose an adaptive model update method controlled by spatio-temporal energy. Considering the model prediction reliability, the spatio-temporal energy $\varepsilon$ is defined as

\begin{equation}\label{varepsilon}
\varepsilon = \underbrace{\frac{y_{max}-\mu_{s}}{\sigma_{s}}}\times \underbrace{\frac{y_{max}-\mu_{t}}{\sigma_{t}}}
\end{equation}
Here, $y_{max}$ denotes the maximum prediction confidence score computed at current frame, $\mu_{s}$ and $\sigma_{s}$ are the mean of and covariance of sidelobe in $M\times N$ score map, respectively. $\mu_{t}$ and $\sigma_{t}$ are the mean of and covariance of $H$ ($H = 5$) previous maximum of prediction scores, respectively. $\varepsilon$ indicates the degree of model reliability, the larger the $\varepsilon$ value, the more stable and reliable the model. $\varepsilon$ is calculated every five frames, assume that $\varepsilon_{0}$ is the spatio-temporal energy value of first five frames. The tracking model is updated, if $\varepsilon \ge \kappa\varepsilon_{0}$ ($\kappa = 0.8$); Otherwise, skip update step. There is a closed-form representation for the gradient of loss $\nabla L(h)$ w.r.t. the network parameter $h$, the model is updated with $h^{i+1} = h^{i}+\alpha \nabla L(h^{i})$. From \cite{DiMP} we can obtain the update rate that follows the steepest gradient direction $\alpha_{s}$, which is derived by minimizing the approximate loss $L$ in the gradient direction,
\begin{equation}\label{SG}
\alpha_{s}= \frac{\nabla L(h^{i})^{T}\nabla L(h^{i})}{\nabla L(h^{i})^{T}\Lambda^{i}\nabla L(h^{i})}
\end{equation}
However, our empirical experiments demonstrate that steepest gradient descend-based update rate $\alpha_{s}$ is too aggressive to maintain model satiability  for some cases. We adopt relatively modest but more efficient update rate $\alpha$ is finalized as
\begin{equation}\label{alpha}
\alpha = \min(\frac{1}{\varepsilon}, \alpha_{s})
\end{equation}
The proposed RSINet tracker is shown in Algorithm \ref{trackingAL}. Since the update process depends on the spatio-temporal energy control, it dynamically update tracking model only with high energy detected frames, and avoid unessential and unreliable update, and the update rate is also modest to resist drastic target appearance change, thus potentially improving the robustness of model and tracking efficiency.

\begin{algorithm}
\SetAlgoLined
\newcommand{\argminD}{\arg\!\min} 
\KwIn{Pre-trained Network model ${\bf M}$ and Initial frame $I_{0}$ with annotation.}
\KwOut{Estimated target state ${\bf \mathcal{O}_{t}^*} = (x_{t}, y_{t}, s_{t}, r_{t})$; \\ \hspace{1.2cm}  Updated model filters $\boldsymbol{h_{t}}$.}
 \While{frame t $\leq$ length(video sequence)}{
 Feed new frame into Siamese network to predict new target state $(x_{t}, y_{t}, s_{t}, r_{t})$.\\
  \If{($t\mid10$)}
  {Calculate spatio-temporal energy $\varepsilon$, in [\ref{varepsilon}]\\
  \If{$\varepsilon \ge \kappa\varepsilon_{0}$}
  {Derive steepest descend update rate $\alpha_{s}$, [\ref{SG}]\\
  $\alpha \gets \min(\frac{1}{\varepsilon}, \alpha_{s})$, [\ref{alpha}]
  }
  Update tracking model filter $h^{t+1} = h^{t}+\alpha \nabla L(h^{t})$.\\
  }
  $t = t + 1$
 }
\caption{Proposed RSINet Tracker.}
\label{trackingAL}
\end{algorithm}

%

\section{Experiments} \label{experiments}
We compare our RSINet tracker with recent state-of-the-art trackers, including ECO \cite{ECO}, SiamRPN \cite{SiameseRPN}, SiamRPN++ \cite{SiamRPN++}, DaSiamRPN \cite{DaSiamese}, TADT \cite{TADT}, ASRCF \cite{ASRCF}, DWSiamese \cite{DWSiamese}, ATOM \cite{ATOM} and DiMP \cite{DiMP}, on the OTB-100 benchmark \cite{OTB-100}, VOT2018 benchmarks \cite{VOT-2018} and LaSOT \cite{LaSOT} dataset, respectively. Some tracking visualization results on OTB-100 and LaSOT datasets are shown as Fig. \ref{fig:tracking_visualization1}, and the quantitative evaluation are presented in Fig. \ref{fig:OTB100_results} and Fig. \ref{fig:LaSOT_results}.
The proposed RSINet tracker is implemented with PyTorch on an Intel Core i7-6700 @3.40-GHz CPU with 24GB RAM and two GTX TITAN X graphics card, running at around 45 frames per second.

\begin{table}[htb]
\newcommand{\tabincell}[2]{\begin{tabular}{@{}#1@{}}#2\end{tabular}}
\small
\centering
  \caption{Ablation analysis of proposed tracker variants. TDD means target-distractor discriminative model, RSI is the rotation-scale invariant model. GD stands for standard gradient descent optimization, SD is steepest descend method proposed in \cite{DiMP}, and AGD is our proposed adaptive gradient descend method.}
  \label{tab:Ablation study}

  \begin{threeparttable}
    \setlength{\tabcolsep}{1.0mm}{
    \begin{spacing}{1.1}
    \begin{tabular}{cccccccc}
    \toprule
    \multirow{2}{*}{}&
    \multicolumn{2}{c}{OTB-100}&\multicolumn{2}{c}{LaSOT}&\multicolumn{3}{c}{VOT2018}\cr
    \cmidrule(lr){2-3} \cmidrule(lr){4-5} \cmidrule(lr){6-8}
    &PR&SR&PR&SR&EAO&A&R\cr
    \midrule
    TDD+GD&     0.802&0.662&0.556&0.388&0.382&0.576&0.186\cr
    TDD+RSI+GD& 0.823&0.678&0.589&0.392&0.411&0.587&0.184\cr
    TDD+RSI+SD& {\bf 0.843}&0.684&{\bf 0.663}&0.556&0.427&0.590&0.176\cr
    \hline
    \tabincell{c}{TDD+RSI+AGD\\({\bf Final Model})}&0.839&{\bf 0.697}&0.660&{\bf 0.585}&{\bf 0.435}&{\bf 0.604}&{\bf 0.143}\cr

    \bottomrule
    \end{tabular}
    \end{spacing}
    }
  \end{threeparttable}
\end{table}

\subsection{Ablation Analysis}
We conduct extensive analysis for proposed RSINet variants on three challenging datasets. Baseline model only applies target-distractor discriminative model with naive gradient descent (TDD+GD) optimizer achieves 0.802 precision score on OTB-100 dataset. By applying proposed rotation-scale invariance model, TDD+RSI+GD gets 0.823 of precision, relatively increasing baseline by 2.6\%, which demonstrates the improtant role of proposed RSI module. When naive gradient descent optimizer is replaced with our proposed adaptive gradient descend method (AGD), the precision score further increases by 4.6\% up to0.839, it further confirms the crucial role of both RSI module and proposed AGD optimizer. The TDD+RSI+SD (steepest descend optimizer) yields the best precision score of 0.843, which means that steepest descend helps more accurately target localization than our final model.

On the other hand, the final model with adaptive gradient descent method approximately outperforms that of steepest descend by 2.0\%, which infers that proposed AGD model can improves overlap accuracy by a considerable margin. The same situation happens on LaSOT benchmark, model with SD is slightly superior than that with AGD (0.4\% precision gain), while AGD based model significantly increases success score by 5.2\% than SD based model. Moreover, the expected average overlap (EAO) EAO (0.435), average accuracy (A) (0.604) and robustness (R) (0.143) of final RSINet tracker outperforms that of baseline model by 14\%, 4.8\%, and 23\%, respectively. Tracking results on three benchmarks prove the obvious effectiveness of proposed modules.

\subsection{Comparison With State-of-the-art Trackers} \label{Benchmark}
{\bf OTB-100 benchmark \cite{OTB-100}:} OTB-100 dataset contains 100 sequences, and evaluates a tracking algorithm for the precision rate and success rate. The precision rate is the average Euclidean distance between the center locations of the tracked targets and the ground-truth bounding boxes. The success rate is a percentage measured by the ratio of successful tracking frames over total frames. Here, a successful tracked frame is the one the overlap score between the tracked bounding box with ground-truth bounding boxes is over 0.5. The success plot shows the ratios of successful frames at overlap thresholds in range [0, 1]. The area under the curve (AUC) of precision and success plot are used to rank the trackers.

\begin{figure}[hbt]
	\centering
	\includegraphics[width= 0.5\textwidth]{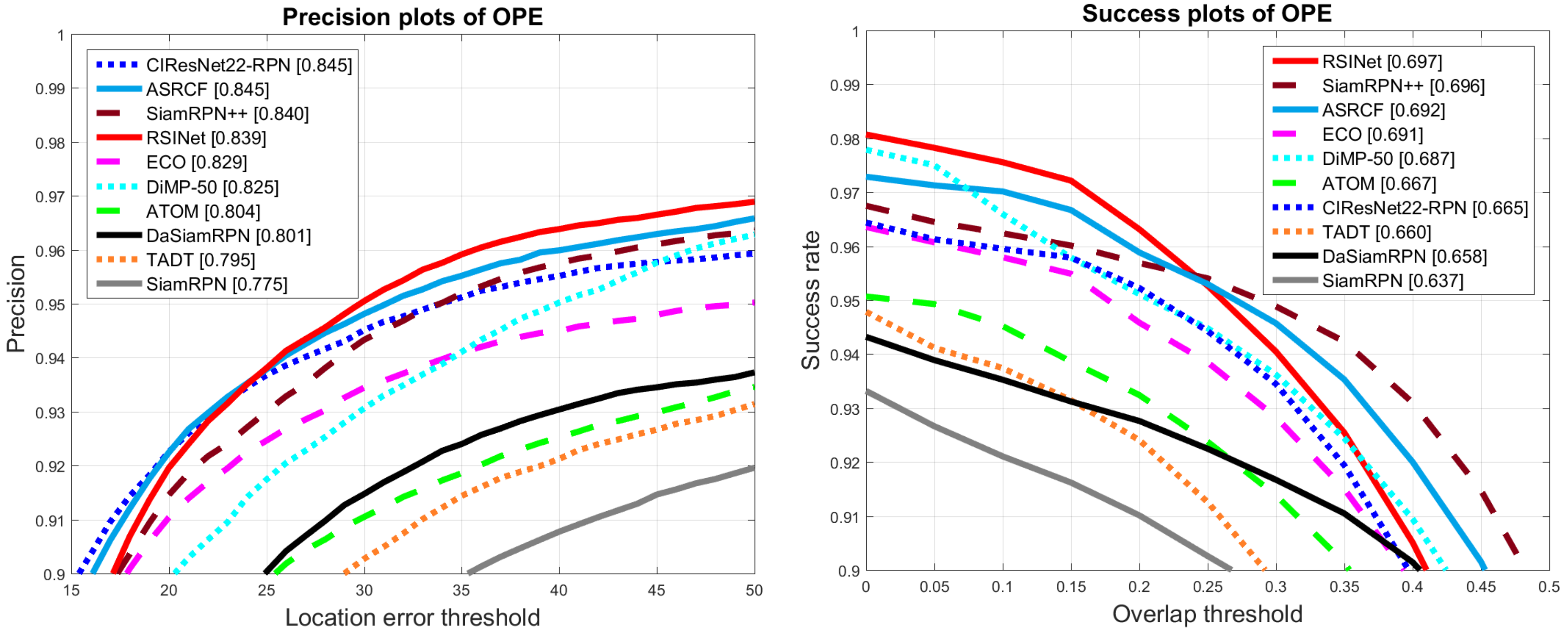}
	\caption{Precision and success plot on the OTB-100 dataset \cite{OTB-100} compared with 8 previous state-of-the-art trackers. Our tracker achieves first place in success rate.}
	\label{fig:OTB100_results}
\end{figure}

We compare our proposed RSINet tracker with recent SOTA trackers tested on OTB-100. The ECO \cite{ECO}  and ASRCF \cite{ASRCF} are the SOTA correlation filter-based trackers integrating both hand-crafted and deep features for target representation. The CIResNet22-RPN is final version of DWSiamese \cite{DWSiamese}, we call DWSiamese in later contents for clarity. DWSiamese \cite{DWSiamese}, SiamRPN \cite{SiameseRPN}, DaSiamRPN \cite{DaSiamese}, and SiamRPN++ \cite{SiamRPN++} are region proposal network-based trackers. The TADT \cite{TADT}, ATOM \cite{ATOM}, and DiMP \cite{DiMP} are deep regression-based trackers, while the later two perform online model update. As shown in Fig. \ref{fig:OTB100_results}, our RSINet achieves 0.829 of precision score, that is comparable result with first place trackers (0.5\% behind). It is duo to the DWSiamese and ASRCF tracker apply both fine-grained and coarse-grained feature maps for position prediction, of which the fine-grained feature maps are very effective for precisely object localization. However, our RSINet and other online deep network based trackers only apply deep feature maps for model representations, thus reducing position discriminative ability. From the respective of success rate, RSINet and most online deep tracker, e.g. ASRCF, ECO, DiMP and ATOM, performs better than DWSiamsese, because that online model update enables trackers timely adopt the target change, including both appearance and scale variance during tracking process, which improves tracking robustness. Our RSINet yields best success tracking performance with 0.697 of success rate.

Table \ref{tab:OTB-100 Attributes results} shows the comparison results between proposed RSINet with other trackers on three representative video attributes, e.g. scale variation (SV), in-plane rotation (IPR) and low resolution (LW). Since RSINet consists of rotation-scale invariant module, we analyze its performance on these three challenging attributes. As shown that RSINet achieves 0.843 of precision rate for scale variation, and 0.856 for in-plane rotation, that outperforms second place tracker (SiamRPN++) by 0.5\% and 1\%, respectively. It is surprising that our tracker achieves 0.664 of success rate for low resolution, that is 3.75\% high than second place tracker (SiamRPN++). Table. \ref{tab:OTB-100 Attributes results} further demonstrates the high effectiveness of proposed RSI module for tackling scale and rotation challenging tracking scenarios.

\begin{table}[htb]
\newcommand{\tabincell}[2]{\begin{tabular}{@{}#1@{}}#2\end{tabular}}
\small
\centering
  \caption{Precision rate and success rate on three representative video attributes on the OTB-100 benchmark, e.g. scale variation (SV), in-plane rotation (IPR), and low resolution (LW).}
  \label{tab:OTB-100 Attributes results}

  \begin{threeparttable}
    \setlength{\tabcolsep}{1.5mm}{
    \begin{spacing}{1.1}
    \begin{tabular}{ccccccc}
    \toprule
    \multirow{2}{*}{Trackers}&
    \multicolumn{3}{c}{precision rate}&\multicolumn{3}{c}{success rate}\cr
    \cmidrule(lr){2-4} \cmidrule(lr){5-7}
    &SV&IPR&LR&SV&IPR&LR\cr
    \midrule
    ECO \cite{ECO}&0.806&0.802&0.804&0.667&0.655&0.603\cr
    DWSiamese \cite{DWSiamese}&0.822&0.835&{\bf 0.826}&0.649&0.662&0.610\cr
    SiamRPN \cite{SiameseRPN}&0.769&0.782&0.795&0.628&0.656&0.597\cr
    SiamRPN++ \cite{SiamRPN++}&0.838&0.848&0.824&{\bf 0.694}&0.694&0.640\cr
    ATOM \cite{ATOM}&0.805&0.794&0.812&0.676&0.650&0.631\cr
    DiMP \cite{DiMP}&0.819&0.835&0.796&0.691&0.686&0.609\cr
    \hline
    {\bf RSINet (Ours)}&{\bf 0.843}&{\bf 0.856}&0.806&0.688&{\bf 0.712}&{\bf 0.664}\cr

    \bottomrule
    \end{tabular}
    \end{spacing}
    }
  \end{threeparttable}
\end{table}

{\bf VOT-2018 benchmark \cite{VOT-2018}:}VOT2018 benchmark contains 60 video sequences, and evaluates the trackers in terms of both robustness (number of failures during tracking), accuracy (average overlap during a period of successful tracking), as well as the expected average overlap (EAO). The accuracy at time-step t measures how well the bounding box predicted by the tracker overlaps the ground-truth bounding box and is defined as the intersection-over-union. The overall accuracy of the $i$-th tracker over a set of $N_{valid}$ valid frames is then calculated as the average of per-frame accuracies. The robustness is defined as the number of times the tracker failed, i.e., the tracking bounding box drifted from the target. The EAO of the VOT challenge is a combination of accuracy and robustness to rank tracking algorithms.

\begin{table}[htb]
\newcommand{\tabincell}[2]{\begin{tabular}{@{}#1@{}}#2\end{tabular}}
\small
\centering
 \caption{Tracking performance of dataset. EAO (Expected Average Overlap) $\uparrow$, A (Accuracy) ranking $\uparrow$ and R (Robustness) ranking $\downarrow$ on VOT2018. {\color{red} First} and {\color{blue} second} performance are marked with red and blue color, respectively.}
 \label{table:VOT2018 tracking results}
 \small
 \setlength{\tabcolsep}{0.7mm}{
 \begin{tabular}{ccccccc}
  \toprule
  & \tabincell{c}{SiamRPN\\\cite{SiameseRPN}}  & \tabincell{c}{SiamRPN++ \\ \cite{SiamRPN++}}& \tabincell{c}{DaSiamRPN\\\cite{DaSiamese}} & \tabincell{c}{ATOM \\\cite{ATOM}}& \tabincell{c}{DiMP \\\cite{DiMP}}&\tabincell{c}{{\bf RSINet}\\({\bf Ours})}\\
  \midrule
  EAO &0.224&0.414&0.383&0.401&{\bf{\color{red}0.440}}&{\bf{\color{blue}0.435}} \\
  A   &0.490&0.600&0.586&0.590&{\bf{\color{blue}0.597}}&{\bf{\color{red}0.604}}\\
  R   &0.460&0.234&0.276&0.204&{\bf{\color{red}0.153}}&{\bf{\color{blue}0.176}} \\
  \bottomrule
 \end{tabular}
 }
\end{table}

\begin{figure}[hbt]
	\centering
	\includegraphics[width= 0.45\textwidth]{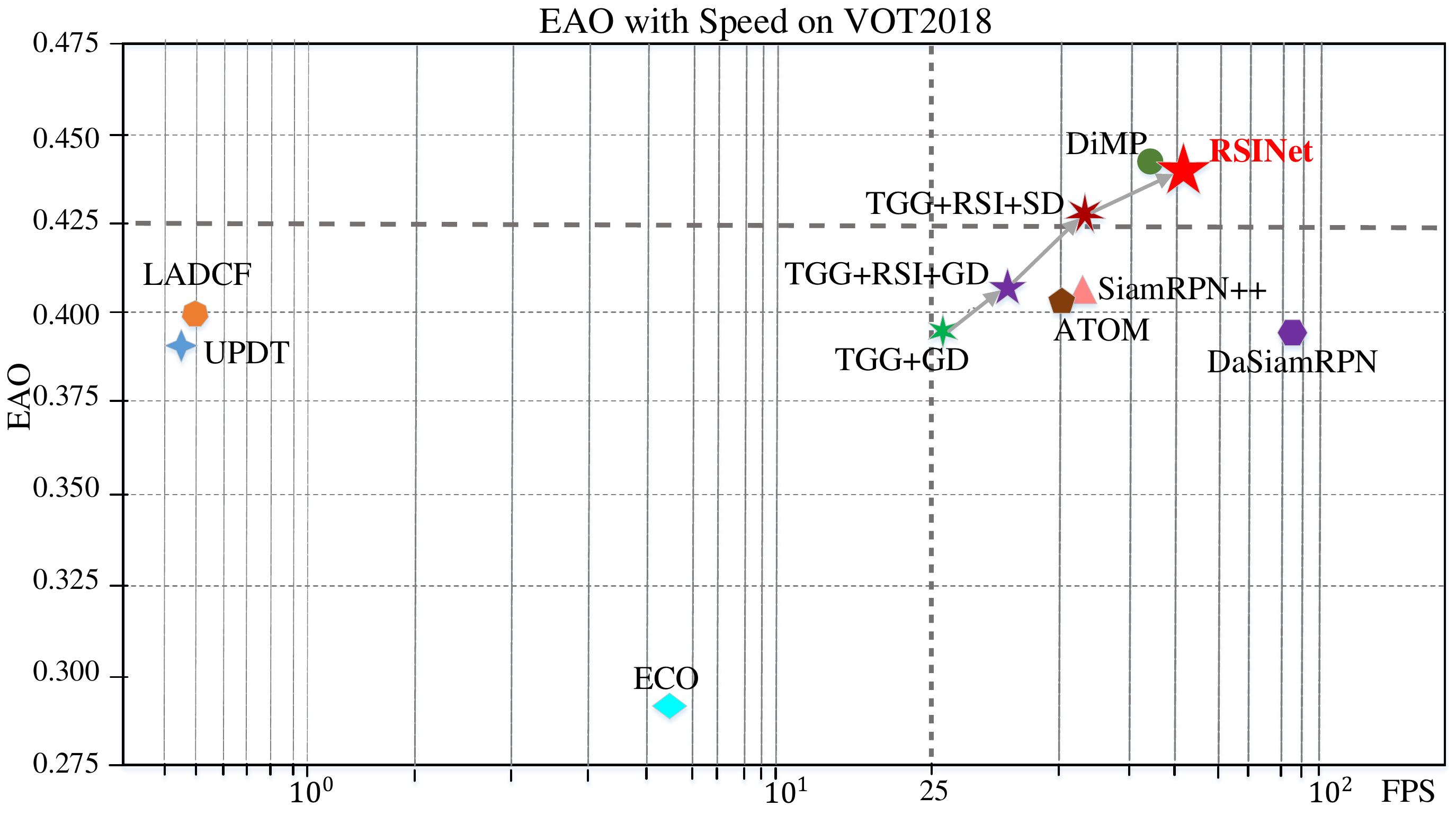}
	\caption{Comparison of EAO (expected aveage overlap) and speed between our proposed RSINet tracker with recent state-of-the-art trackers on VOT2018 benchmark. The x-axis is the speed value (frame per second) in log space. TDD+RSI+AGD is our final RSINet tracking model, and TDD+X(+X) is its variants.}
	\label{fig:VOT2018 results}
\end{figure}

We evaluate our RSINet tracker on VOT2018, and compared with region propose network(RPN)-based trackers, e.g. SiamRPN \cite{SiameseRPN}, DaSiamRPN \cite{DaSiamese}, and SiamRPN++ \cite{SiamRPN++}, as well as deep regression-based trackers, ATOM \cite{ATOM} and DiMP \cite{DiMP}. As shown in Table \ref{table:VOT2018 tracking results}, our RSINet are beyond RPN-based trackers for all evaluation metrics, e.g. 0.435 of EAO, 0.604 of A and 0.176 of R. And RSINet tracker achieves remarkable 0.604 of accuracy, yielding a new state-of-the-art on VOT2018 benchmark, exceeding DiMP by 1.17\%. From the third row of Table \ref{table:VOT2018 tracking results}, we can see that our RSINet tracker is less robust than DiMP, we analyze that our tracker need to be improved target re-identifying capability in long-term tracking challenge. We address this issue for our future works. Fig. \ref{fig:VOT2018 results} gives a visible comparison with respect of EAO and speed performance, our RSINet maintains a good balance between the accuracy and speed, which is a remarkable advantage for real applications.

\begin{figure}[hbt]
	\centering
	\includegraphics[width= 0.5\textwidth]{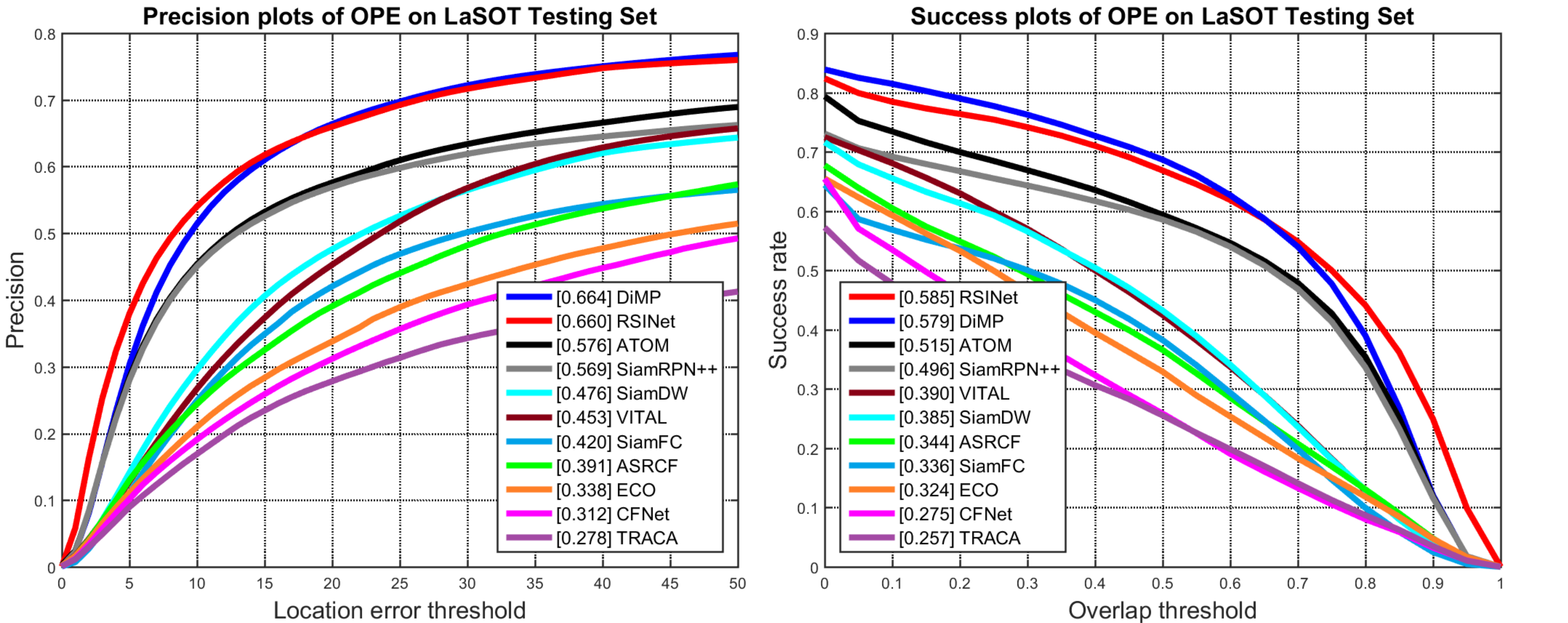}
	\caption{Precision and success plot on LaSOT dataset. Our RSINet achieves second place for precision rate and outperforms all previous trackers.}
	\label{fig:LaSOT_results}
\end{figure}

{\bf LaSOT benchmark \cite{LaSOT}} LaSOT benchmark contains 1400 sequences for training and testing, we evaluate our RSINet tracker on 280 testing sequences. The precision is computed by measuring the distance between tracked result and the ground-truth bounding box in pixes. Trakcers are ranked with this meteric (20 pixels)and using the area under the curve (AUC) in range [0, 0.5]. The success is computed as the intersection over union (IoU) between tracked and ground-truth bounding box, the trackers are ranked using the AUC in range [0, 1].
We perform experiments on recent large-scale LaSOT benchmark, which is suitable for testing the generalization ability of trackers. Fig. \ref{fig:LaSOT_results} depicts that our RSINet is very competitive compared to other SOTA trakers in both precision rate and success rate evaluation. Specifically, our tracker achieves 0.664 of precision, 0.6\% lower than champion tracker (DiMP). However, RSINet gets 0.585 of success rate that slightly outperforms DiMP by 1.0\%. We attribute this good performance to the online stable model update strategy.


\section{Conclusion}\label{conclusion}

We propose a rotation-scale invariant tracking framework, RSINet, that enables target-distractor model and rotation-scale model learning simultaneously. RSINet consists of two branches, TDD branch and RSI branch. TDD branch learns target discriminative model, while RSI model learn rotation and scale changes during tracking. For model stability and reliability, we propose a simple yet efficient model update method, which potentially improves tracking robustness and tracking speed. Experiments on three challenging tracking benchmark demonstrate proposed RSINet achieves comparable performance compared to recent state-of-the-art trackers, while RSINet maintains a good balance between tracking accuracy (0.604 on VOT2018) and running efficiency (45 FPS). And we address the drawback of our tracker in long-term tracking challenge for future works.

\section*{Acknowledgments}
This research was partially supported by the Ministry of Science and ICT (MSIT), Korea, under the Information Technology Research Center (ITRC) support program (IITP-2017-0-01642). This research was also partly supported by project under Grant WQ20165000357 of Chinese Government.


\end{document}